%% file: main.tex
\documentclass[runningheads]{llncs}

\usepackage{eccv}

\usepackage{eccvabbrv}

\usepackage{graphicx}
\usepackage{booktabs}
\usepackage{multirow} 
\usepackage{amssymb}
\usepackage{colortbl} 
\usepackage{tabularx,multirow}
\usepackage[colorlinks=true, linkcolor=red, urlcolor=red]{hyperref}

\usepackage[accsupp]{axessibility}  % Improves PDF readability for those with disabilities.

\usepackage{hyperref}

\usepackage{orcidlink}

\begin{document}

\title{Animate Your Motion: Turning Still Images into Dynamic Videos} 

\titlerunning{Animate Your Motion}

\author{ Mingxiao Li$^*$\inst{1} \and
 Bo Wan$^*$\inst{2} \and
Marie-Francine Moens\inst{1} \and 
Tinne Tuytelaars\inst{2}}

\authorrunning{Li et al.}

\institute{Department of Computer Science, KU Leuven
\email{\{mingxiao.li,sien.moens\}@kuleuven.be}\\
%\url{http://www.springer.com/gp/computer-science/lncs} 
\and
Department of Electrical Engineering, KU Leuven\\
\email{\{bo.wan,Tinne.Tuytelaars\}@esat.kuleuven.be} \\
}

\maketitle
\def\thefootnote{*}\footnotetext{Equal Contribution. Project page: \href{https://mingxiao-li.github.io/smcd/}{https://mingxiao-li.github.io/smcd/}}\def\thefootnote{\arabic{footnote}}

\begin{figure}[ht]
    \centering
    \includegraphics[width=1.0\linewidth]{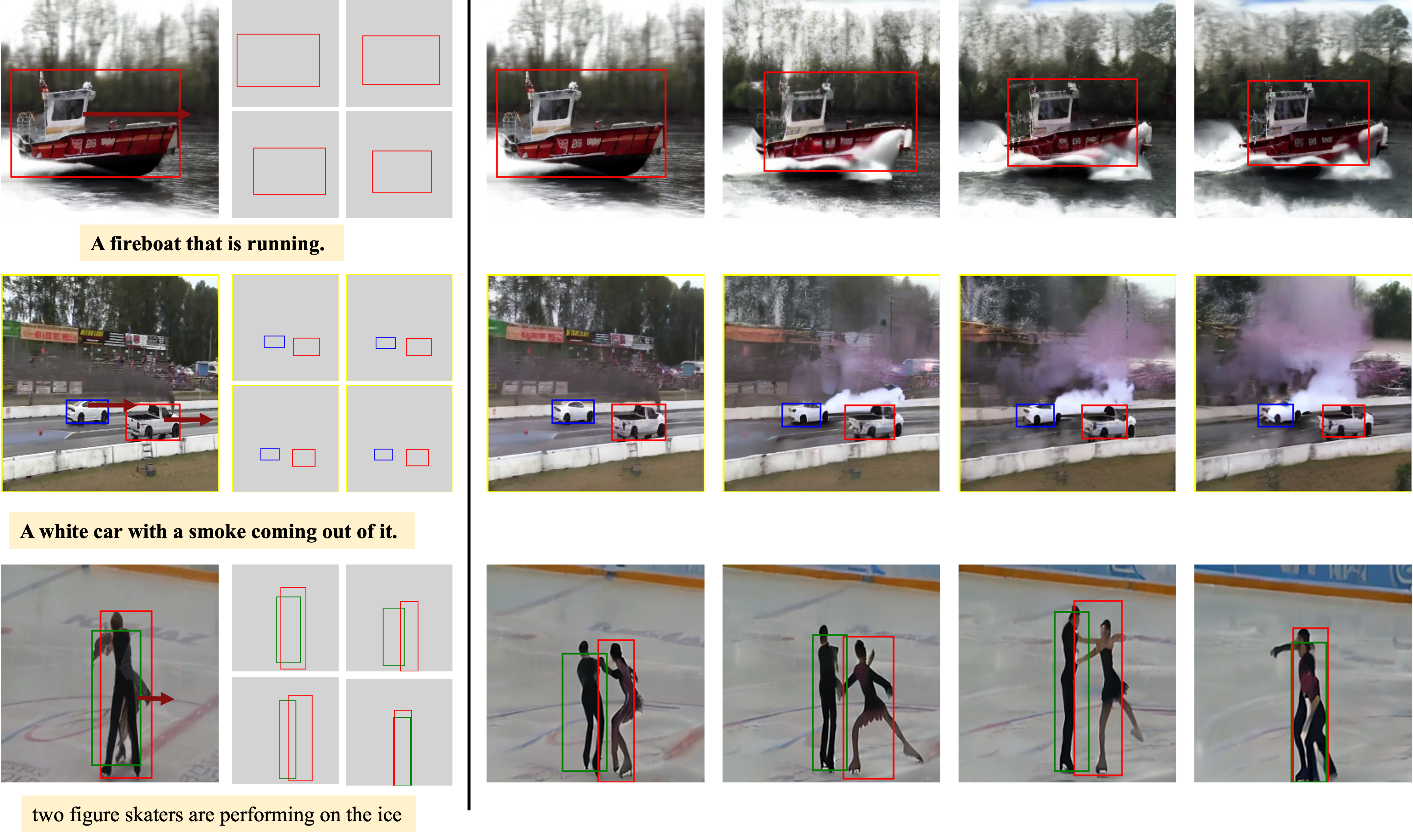}
    \caption{\textbf{Scene motion customized video generation results} of our proposed model. Our model accepts an initial frame image, a sequence of bounding boxes, and text as inputs to generate the desired videos that comply with the given constraints. The red arrows in the image indicate the moving directions of the objects.}
    \label{fig:visual-1}
\end{figure}
\label{sec:intro}

\begin{abstract}
In recent years, diffusion models have made remarkable strides in text-to-video generation, sparking a quest for enhanced control over video outputs to more accurately reflect user intentions. 
Traditional efforts predominantly focus on employing either semantic cues, like images or depth maps, or motion-based conditions, like moving sketches or object bounding boxes.
Semantic inputs offer a rich scene context but lack detailed motion specificity; conversely, motion inputs provide precise trajectory information but miss the broader semantic narrative.
For the first time, we integrate both semantic and motion cues within a diffusion model for video generation, as demonstrated in Fig.~\ref{fig:visual-1}.
To this end, we introduce the Scene and Motion Conditional Diffusion (SMCD), a novel methodology for managing multimodal inputs. It incorporates a recognized motion conditioning module~\cite{li2023gligen} and investigates various approaches to integrate scene conditions, promoting synergy between different modalities. 
For model training, we separate the conditions for the two modalities, introducing a two-stage training pipeline.
Experimental results demonstrate that our design significantly enhances video quality, motion precision, and semantic coherence.
\end{abstract}

\section{Introduction}
% background: image/video generation
Diffusion models have ushered in a new era of high-quality text-to-image (T2I) generation, providing a robust and scalable framework for training. These models, celebrated for their ability to generate detailed and coherent images from textual descriptions, have significantly advanced the field of generative models. Inspired by the success in T2I generation, researchers have begun to explore the extension of these models' capabilities to video content. This has led to the development of innovative approaches in text-to-video (T2V) generation~\cite{wang2023modelscope,makevideo,zhou2022magicvideo} and editing~\cite{kahatapitiya2024object,li2023vidtome,deng2023dragvideo}, which promise to revolutionize how dynamic visual content is created and manipulated. 

% existing works: conditional video generation
Among the groundbreaking contributions to this field, one particularly notable work is Sora~\cite{liu2024sora}, which leverages the diffusion transformer structure. This structure is renowned for its exceptional scalability across diverse domains, enabling it to generate up to a minute of high-fidelity video. 
Despite these advances, T2V generation is confronted with a fundamental challenge: the output space for videos is exponentially larger than that for images, making accurate generation more complex. This complexity necessitates additional constraints to ensure that the generated content aligns more closely with user expectations. Addressing this challenge, recent studies have introduced various forms of conditional inputs to refine video generation. These enhancements include semantic conditions, such as incorporating static images~\cite{girdhar2023emu} and depth information~\cite{wang2024videocomposer}, as well as motion-based conditions, like moving sketches and object bounding boxes~\cite{tracking}, to guide the generation process more effectively.

% our task
While the aforementioned studies have made significant strides in enhancing T2V generation, they predominantly focus on leveraging either semantic or motion conditions in isolation. This singular approach misses the opportunity to combine the strength of both semantic and motion cues for a more controlled and customized video generation. Recognizing this gap, our work introduces an innovative framework that integrates both semantic imagery and motion trajectories as inputs. As illustrated in Fig.~\ref{fig:visual-1}, our objective is to generate videos where objects not only exhibit motion but do so in a manner that is semantically consistent with the provided image. To our knowledge, this represents the first attempt to synergize control signals from these two distinct domains, thereby offering a novel way to animate static images with custom motions. This dual-input approach promises a stronger regularization for video generation, opening new avenues for creating highly customized and dynamic visual content.

% task challenges
In this task, we start with an image containing multiple objects and their movement trajectories, each depicted as a sequence of bounding boxes and its semantic category. Our goal is to generate the subsequent video frames that align with these trajectories, thereby animating the image in a way that is both flexible and interactive.
Despite the compelling nature of our objective, we encounter several practical challenges. 
% conv / controlnet / cross-attn
Firstly, integrating diverse signals as conditional inputs into video generation is a complex task that requires careful consideration of various strategies. We delve into this issue in our ablation studies, where we explore and compare different methods of signal integration.
% combination of conv+cross-attn
Second, ensuring consistency between the moving objects and the static background across frames is challenging but crucial for realism.
% stage-wise tuning
Lastly, given that these signals originate from fundamentally different domains, finding a balance that accommodates their respective requirements without compromising the overall quality of the generated video is a delicate task.

% our method
To address the aforementioned challenges, we introduce a novel Scene and Motion Conditional Diffusion model (SMCD), which innovates on top of a pretrained T2V diffusion model, typically based on a UNet or DiT architecture. To this foundation, we add two specialized modules designed to process conditional image and motion inputs separately.
Specifically, the core of SMCD is a parallel video generation pipeline that predicts all video frames simultaneously. Drawing inspiration from~\cite{li2023gligen}, SMCD includes a motion integration module (MIM) that utilizes a gated self-attention layer within each UNet block to encode the box locations to the visual tokens. 
Additionally, SMCD adopts a dual image integration module (DIIM) comprising a zero-convolutional layer, as used in ControlNet~\cite{zhang2023adding}, and a gated cross-attention layer. 
The zero-convolutional layer is incorporated with the image features to progressively modulate the video content, complemented by a gated cross-attention layer designed to balance the influence of the image condition across each UNet block.
This dual approach allows SMCD to impose image conditions across the entire generation process, enhancing the coherence and consistency of the generated frames.
Interestingly, we find that simultaneously training these two signal integration modules can lead to competitive interference, resulting in outputs of inferior quality. To mitigate this, we propose a two-stage training strategy. We first train the motion integration module and then, with this module frozen, proceed to train the dual image integration module. This sequential approach prevents the competition between signals, leading to the generation of cleaner and more focused videos.

% conclusion
We summarize our contributions as follows: (i) We introduce a novel task in the realm of video generation, focused on conditioning the process with both imagery and motion, enhancing user interactivity and customization; (ii) We propose SMCD, which innovatively integrates diverse input signals to guide the generation process; (iii) We demonstrate SMCD's ability to produce videos in which objects not only exhibit a predefined motion but do so in a manner that is semantically consistent with the provided image.

\section{Related Works}
\label{sec:rw}
\subsubsection{Diffusion Models}
Diffusion Models, as introduced by~\cite{ddpm} and inspired by nonequilibrium thermodynamics, are probabilistic frameworks capable of transforming data into Gaussian noise through a forward process and subsequently reverting Gaussian noise back to the data distribution in a reverse process. These models have demonstrated superior performance in content generation tasks, including text-to-image conversion~\cite{stable_diffusion}, 3D object creation~\cite{poole2022dreamfusion}, and audio production~\cite{chen2020wavegrad}.The inherent iterative process of these models, which typically requires hundreds of steps to generate content, has seen significant efficiency improvements. \cite{ddim} introduced the Denoising Diffusion Implicit Model (DDIM), which decreases the number of necessary steps for generating high-quality outputs. Further advancements such as the application of ordinary differential equation solvers~\cite{dpmsolver,dpmsolver2,pseudo}, optimization of variance~\cite{bao2022analytic}, reduction of exposure bias~\cite{li2023alleviating,ning2023elucidating,ning2023input}, and enhancements in noise scheduling~\cite{improved_ddpm} have all contributed to faster inference and enhanced generative performance.

\subsubsection{Video Generation with Diffusion Models}
The initial progress in generating videos with Diffusion Models can largely be credited to the pioneering work of~\cite{vdm}, who introduced the concept of a 3D diffusion UNet. This was followed by an innovative approach by~\cite{imagen}, which combined a cascading sampling framework with super-resolution techniques to produce high-quality videos. Further advancements were made with the introduction of a temporal attention mechanism~\cite{makevideo}, enhancing frame-by-frame coherence. Additionally, MagicVideo~\cite{zhou2022magicvideo} and LVDM~\cite{lvdm} integrated this mechanism into latent Diffusion Models for video generation. In our research, we employ the ModelScope text-to-video generation model~\cite{wang2023modelscope} as our backbone. 

\subsubsection{Customized Generation} Customized generation, in contrast to general generation tasks, is more adept at aligning with user preferences. The majority of research focuses on customized image generation. This involves creating images based on various controls, including edges, depth, brain activity, etc., as demonstrated in studies by~\cite{zhang2023adding,li2023gligen,sun2024contrast,sun2024neurocine,sun2023decoding}. Additionally, there is a significant interest in subject personalized image generation, as explored by~\cite{ruiz2023dreambooth,tinver}. Recent studies, including~\cite{videocontrol,zhang2023controlvideo}, have expanded the scope of conditional image generation models to include video generation, which is conditioned on various sequences like edges, depth, and segmentation. Concurrently, another body of research ~\cite{ma2023trailblazer,chen2024motion,wang2024boximator,wang2024videocomposer} has been dedicated to developing video generation models that allow for motion control. Our work diverges significantly from these existing studies by introducing, for the first time, a model for scene, motion and text conditional video generation. Our model is unique in its ability to generate videos based on a combination of a user-provided image, a motion sequence, and a textual description. The image contributes detailed semantics; the motion sequence accurately describes object dynamics; and the text specifies the precise motion state. Utilizing these three control signals, our model is capable of producing videos that are more closely aligned with user preferences.

\begin{figure}[t]
    \centering
    \includegraphics[width=01.0\linewidth]{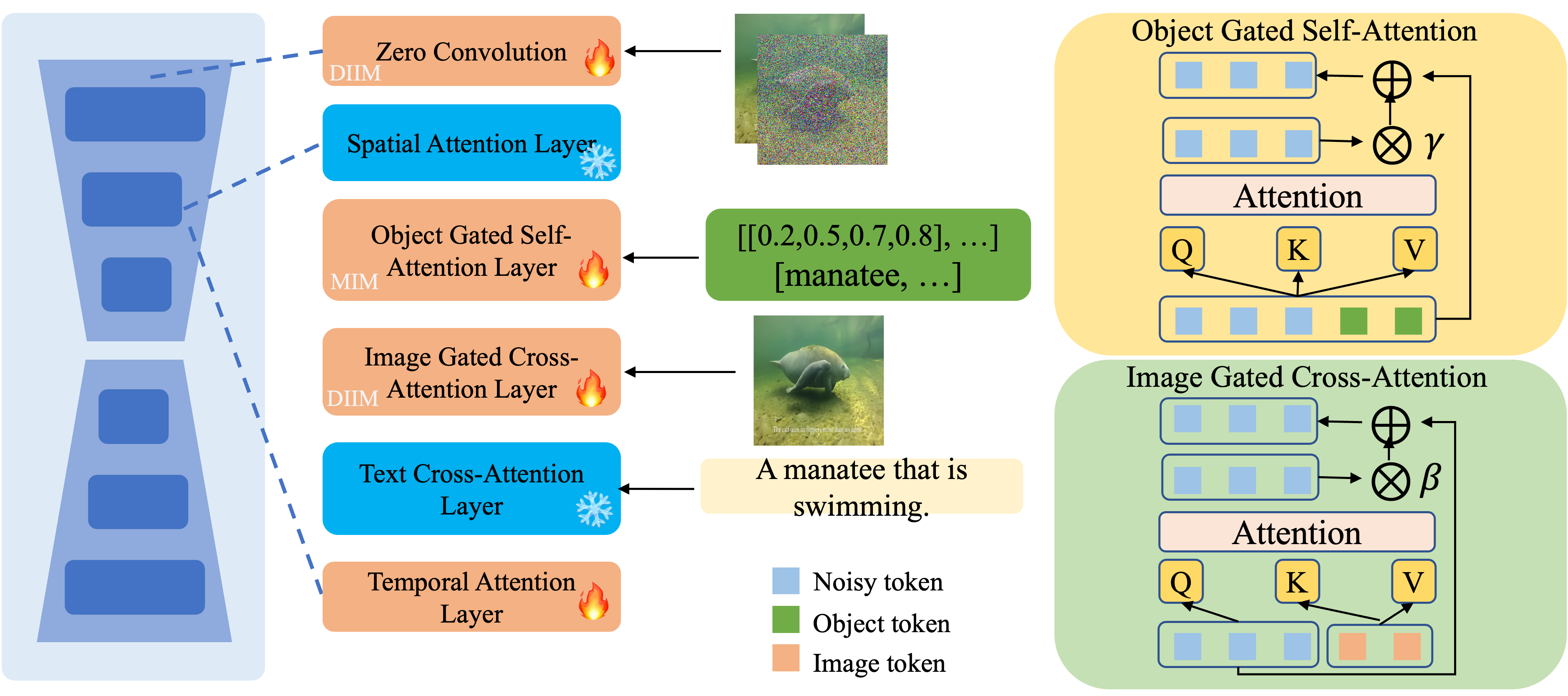}
    \caption{\textbf{Model Illustration}: SMCD handles three control signals including images, 
    bounding box sequences, and text. It builds on a pre-trained T2V model, enriched with an object-gated self-attention layer, 
    image-gated cross-attention layer, and a zero initialized convolution layer. These enhancements allow it to adapt to bounding box and image conditions. 
    % through a two-stage training process: first focusing on the object-gated self-attention, followed by the zero convolution and image-gated cross-attention layers.
    }
    \label{fig:model}
\end{figure}

\section{Method}
\label{sec:method}

Before we introduce our Scene and Motion Conditional Diffusion Model in section~\ref{sec:SMCD}, we first give a precise definition of the task we are addressing in this work (section~\ref{sec:task}) and recap the working of the baseline model (section~\ref{sec:prelim}). 

\subsection{Task Definition}
\label{sec:task}
In this work, our goal is to transform still images into dynamic videos 
with customized object movements. Formally, given an image frame $v_0$ and 
a set of object trajectories $O=\{o_1,o_2,\cdots,o_N\}$, where $o_i=\{(b_{i,1}, b_{i,2}, \cdots b_{i,F}), c_{i}\}$ represents the trajectory and category of the $i$-th object across $F$ frames, with $N$ representing the total number of objects. Here $b_{i,f}$ denotes the bounding box coordinates of the $i-$th instance in $f$-th frame, and $c_{i}$ is its textual label. Our objective is to generate a video sequence $V = \{v_1, v_2, \cdots, v_F\}$, which should accurately reflect the motion as delineated by the object trajectories $O$, while maintaining semantic integrity and the details of the given image condition $v_0$.

\subsection{Base model}
\label{sec:prelim}
\subsubsection{Image generation with Latent Diffusion Models} Latent Diffusion Models (LDMs)~\cite{stable_diffusion} have demonstrated exceptional performance in generating images from text prompts. The forward training process for LDMs begins by encoding an image $x$ into a latent representation $z_0=\mathcal{E}(x)$ using a pretrained VAE encoder. This latent representation is progressively transformed into Gaussian noise through the process $z_t = \sqrt{\bar{\alpha}}_t z_0 + \sqrt{1-\bar{\alpha}}_t \epsilon$, where $\epsilon$ is noise sampled from a standard normal distribution ($\epsilon \sim \mathcal{N}(0,1)$), and $\bar{\alpha}_t$ represents a predetermined noise schedule. The training objective is to precisely predict the noise added at each timestep, with a loss function defined as $\mathcal{L}^{simple} = \mathbb{E}_{t,\epsilon_t \sim \mathcal{N}(0,1)}[|\epsilon_t - \epsilon_{\theta}(z_t, c)|_2^2]$, where $t$ denotes the diffusion timestep, and $c$ is the conditioning text prompt. In the reverse generation phase, LDMs gradually reconstruct the image features in alignment with the text prompt from Gaussian noise. The denoised output is then passed through the VAE decoder to recreate the full-color image.
% : $x_0 = \mathcal{D}(z_0)$
%

\subsubsection{Video generation with ModelScope} Recent advancements have extended LDMs to video generation, applying a similar approach but with modifications to accommodate video inputs. Many of these developments build on the ModelScope~\cite{wang2023modelscope} architecture, which utilizes a 3D diffusion UNet designed to process temporal information, as explained below. 
%. Distinguishing itself from the conventional 2D UNet~\cite{stable_diffusion}, ModelScope introduces both a 3D temporal convolution and a temporal attention layer within each block of the 3D UNet, crucial for preserving the temporal coherence of the generated videos.

Given a video data sample $V$, we follow the LDM~\cite{stable_diffusion} 
and leverage a pretrained VAE~\cite{kingma2013auto} to project each frame $v_f$ in video $V$ from pixel space to latent space: 
$x_f=\mathcal{E}(v_f)$, where $x_f \in \mathbb{R}^{C \times H \times W}$ represents the latent representation of the $f$-th frame, with $C,H,W$ indicating the channel, height, and width dimensions, respectively.
The collective latent representations of the video frames are represented as $Z_0 = \{x_1, x_2, \cdots, x_F\}$, 
where $Z_0 \in \mathbb{R}^{F \times C \times H \times W}$ encapsulates the entire video's latent space.
% Notably, the first frame's latent representation $x_1$ is designated as the image condition for the subsequent video generation process.
%
% Additionally, for the initial frame of the video, we obtain its latent representation $x_{0} \in \mathbb{R}^{C \times H \times W}$ using the same VAE. 
Further, the given image caption $c$ and the object labels $\{c_i\}$ are preprocessed using a pretrained CLIP textual encoder $f_{\text{text}}$.
%To enhance the understanding of the semantic content, we employ the pretrained CLIP textual encoder $f_{\text{text}}$ to encode the caption $c$.
% and the object labels $\{c_i\}$.

%\subsubsection{3D Diffusion UNet} 
The diffusion process gradually introduces Gaussian noise to the video features $Z_0$, resulting in a series of noised video features $\{Z_t\}_{t=1}^T$. Video diffusion models (VDMs) are then tasked with reversing this process, aiming to recover the original video features $Z_0$ from the terminal Gaussian noise state $Z_T$. 
For simplification, when discussing the denoising process, we will consider a single timestep $t$ as a representative example and momentarily disregard the timestep subscript for the noised video features $Z_t$. Consequently, the input for a VDM is denoted as $Z \in \mathbb{R}^{F \times C \times H \times W}$.

% a ResNet block
A 3D diffusion UNet~\cite{wang2023modelscope}, recognized for its effectiveness in VDMs, consists of three critical components: a spatial attention layer, a text cross-attention layer, and a temporal attention layer. 
% The process begins with a ResNet equipped with 3D convolution layers, tasked with capturing semantic information across both spatial and temporal dimensions.
% \begin{equation}
% Z = \text{ResNet}(Z)
% \label{eq:resnet}
% \end{equation}
The process begins with a spatial-wise self-attention employed on the visual tokens within each frame to enhance the encoding of visual context:
\begin{equation}
z_f = z_f + \text{SelfAttn}(z_f)
\label{eq:self-attn}
\end{equation}
where $z_f\in \mathbb{R}^{C \times H \times W}$ represents the noised image features of the $f$-th frame in $Z$. Following this, a cross-attention layer integrates textual semantics from the image caption for each frame:
%for a comprehensive understanding of both visual and textual information:
\begin{equation}
z_f = z_f + \text{CrossAttn}(z_f, f_{\text{text}}(c))
\label{eq:cross-attn}
\end{equation}
Finally, the temporal attention layer consolidates context across different frames, ensuring coherent temporal progression throughout the video sequence:
\begin{equation}
Z = Z + \text{TempAttn}(Z)
\end{equation}

\subsection{SMCD}
\label{sec:SMCD}
Our proposed model, the Scene and Motion Conditional Diffusion Model (SMCD), evolves from ModelScope~\cite{wang2023modelscope} and extends it by incorporating both object trajectories and the initial image frame as pivotal conditional inputs. This advancement is facilitated through the introduction of two specialized modules: a {\em motion integration module} (MIM) and a {\em dual image integration module} (DIIM), as showcased in Fig~\ref{fig:model}.
The motion integration module is designed to precisely capture and interpret the dynamic trajectories of objects, while the dual image integration module focuses on retaining the intricate semantic details inherent in the initial image frame. Together, these modules form a comprehensive system that adeptly marries dynamic motion information with static semantic imagery.

\subsubsection{Motion Integration Module}
Inspired by GLIGEN~\cite{li2023gligen}, we enhance SMCD by freezing the pretrained 3D diffusion UNet from ModelScope and introducing a motion integration module to encode object trajectories. Specifically, within each frame, we start by generating box location tokens \(s_{i,f}\) through a multi-layer perceptron (MLP), which involves encoding the bounding box coordinates \(b_{i,f}\) and their associated labels \(c_i\):
\begin{equation}
s_{i,f} = \text{MLP}(\text{Fourier}(b_{i,f}), f_{\text{text}}(c_i))
\end{equation}
here Fourier embedding~\cite{tancik2020fourfeat} is employed to process box coordinates \(b_{i,f}\) and the object category label $c_i$ is preprocessed by the same pretrained CLIP textual encoder used for encoding the image caption.

Subsequently, a gated self-attention layer is integrated in each UNet block, positioned between the spatial attention layer (Eq.~\ref{eq:self-attn}) and the text cross-attention layer (Eq.~\ref{eq:cross-attn}). It enables the model to concentrate on the moving objects by performing self-attention over the concatenation of visual and location tokens
% %$[z_f, s_{i,f}]$:
% $[z_f, s_{1,f}, ..., s_{N,f}
% ]$
:
\begin{equation}
    z_f = z_f + \tanh(\gamma) \cdot TS(\text{SelfAttn}([z_f, %s_{i,f}]))
s_{1,f}, ..., s_{N,f}]))
\end{equation}
where TS(·) is a token selection operation that filters for visual tokens, and \(\gamma\) is a trainable scalar. Initially set to zero, \(\gamma\) is tuned to calibrate the influence of location cues. For further details, we refer to~\cite{li2023gligen}.

\subsubsection{Dual Image Integration Module} 
% Thus we found some inconsistency with the condition images for some cases, as validated in Sec. xxx.
% Previous works [cite] validated simply adding the
% the image feature as residual to the Gaussian noise and input it to the first block of UNet could achieve such a goal.
% However, the image signals become weaker when they go to deeper blocks, 
% even it has residual connections but loses information in the feature compression.

To enhance the video generation process with image conditions, we introduce a dual image integration module that enriches the video features with image context via zero-conv (ZC) and further enhances this through a gated cross-attention (GCA) mechanism.

Firstly, given the conditional image $v_0$, a convolutional network is adopted to process its latent feature $x_0=\mathcal{E}(v_0)$, with parameters initialized to zero. This is inspired by \cite{zhang2023adding}, which allows $x_0$ to gradually influence the video content as the model learns.
The processed image feature is combined with the video features before they are input into the 3D UNet:
\begin{equation}
    z_f = z_f + \text{ZeroConv}(x_0)
    \label{eq:zero_conv}
\end{equation}

Additionally a gated cross-attention is applied to further amplify the image condition's impact across each UNet block. 
Here, $x_0$ acts as both the key and value, with the video frame features $z_f$ serving as the query:
\begin{equation}
    \hat{x}_0 = \text{ResNet}(x_0); \quad
    z_f = z_f + \tanh(\beta) \cdot \text{CrossAttn}(q(z_f), k(\hat{x}_0), v(\hat{x}_0))
\label{eq:cross_gated_attn}
\end{equation}
where an extra ResNet layer adapts $x_0$ into the feature space $z_f$, and $\beta$ is a trainable scalar initialized as zero to control the influence of image signals.
% thus Eq.~\ref{eq:resnet} in the first UNet block can be re-written as

% ControNet
% we use a copied encoder to deal with the image feature input $\hat{x}_0$, 
% and these features are input to the corresponding output blocks by concatenation

\subsection{Model Training} ~\label{sec:model_train}
SMCD is initialized from pretrained Modelscope. During model training, we freeze the spatial attention layers and text cross-attention layers to preserve the integrity of the model's initial capabilities,
and train the remaining parts in a two-stage pipeline. 
In the first stage, the primary objective is to refine the model's proficiency in managing object locations within a single image. To achieve this, we train the motion integration module on image frames without considering any temporal information. 
In the second stage, we enhance the model with the capability to consider the image condition as well as temporal coherence. During this phase, we train the parameters of the dual image integration module along with the temporal attention layer of the 3D UNet. This training is conducted on the video dataset.
For both stages, we adopt classifier-free guidance training. 
Specifically, in the first stage of training, box information was randomly omitted with a probability of $p_b$. In the subsequent stage, both the image and box sequence were randomly excluded, with probabilities $p_i$ and $p_b$, respectively.
This strategy introduces an element of unpredictability that encourages the model to learn robust feature representations.
Inspired by~\cite{ddpm}, we optimize the parameters $\theta$ in SMCD with a similar training objective:
\begin{equation}
    \mathcal{L} = \mathbb{E}_{t,\epsilon_t \sim \mathcal{N}(0,1)}[|\epsilon_t - \epsilon_{\theta}(Z_t, c, v_0, O)|_2^2]
\end{equation}

\subsection{Inference}
In the inference stage, we sample a Gaussian noise $Z_T$ and run the denoising process to generate the clean video latent representation $Z_0$. Following this, the latent representation is transformed into an RGB video using the VAE decoder. Our model is designed to create videos based on three specific conditions: text prompt $c$, the conditional image $v_0$, and a sequence of box trajectories $O$. 
To enhance its capability, we apply the classifier-free guidance approach during inference, which can be formulated as follows with a guidance scalar $\alpha$:
\begin{equation}
    \hat{\varepsilon}= \varepsilon_{\theta}(Z_t,c,v_0,O) + \alpha \cdot (\varepsilon_{\theta}(Z_t,c,v_0,O) - \varepsilon_{\theta}(Z_t,\emptyset,v_0,O))
\end{equation}

\section{Experiments}
\label{sec:exp}

\subsection{Experimental Setup}
\subsubsection{Dataset}
Our research utilizes the publicly available GOT10K~\cite{huang2019got} and YTVIS 2021~\cite{yang20224th} datasets, both of which feature sequences of object bounding box annotations. The GOT10K dataset is widely used in single object tracking research, comprising more than 10,000 video segments dedicated to tracking real-world moving objects. Conversely, the YTVIS2021 dataset is a foundational resource for video instance segmentation and multi-object tracking research. It contains 2,985 training videos annotated with high-quality bounding boxes across 40 semantic categories. 
We divided YTVIS into 2,852 train and 160 test videos, following the split in \cite{yang20224th}. For GOT10K, we used the standard split, resulting in 9,335 train and 180 validation videos.
Similar to the work of~\cite{li2023trackdiffusion}, we employ the LLaVA~\cite{liu2023improved} model to generate captions for both datasets.

\subsubsection{Training details}
We adopt a two-stage training pipeline.
In the first phase of training, we extract frames from the videos in GOT10K and YTVIS2021 datasets and create data samples composed of images along with object boxes. The model is trained for approximately 50k steps with a batch size of 32. In the second phase, we continue to fine-tune the model on the videos, extending the training for an additional 80k steps with a reduced batch size of 8. 
The probabilities for randomly excluding images and bounding boxes are set at $p_i=0.25$ and $p_b=0.1$, respectively.
We use AdamW optimizer~\cite{loshchilov2017decoupled} with a learning rate of $5e^{-5}$. All our experiments are conducted using a single NVIDIA A100 GPU. Each experiment needs three days to finish training. 

\subsubsection{Model details}  
We generate videos with a resolution of 256$^2$ pixels, producing $F=8$ frames simultaneously. Given that the VAE encoder reduces the image resolution by a factor of 8, the width $W$ and height $H$ of the latent feature in our method are set to 32. For further information on the CLIP text encoder, VAE models, and the 3D diffusion UNet, please refer to ModelScope~\cite{wang2023modelscope}.

\subsubsection{Evaluation Metrics}
In our analysis, we evaluate SMCD on the validation sets of the GOT10K and YTVIS2021 datasets. Following the previous work, we use FVD~\cite{fvd} to evaluate video quality. Additionally, we leverage the pretrained CLIP model~\cite{clip} to compute the CLIP text-image similarity score (CLIP-SIM) for individual frames in each video. To evaluate the First Frame Fidelity (FFF), a metric that measures the similarity between the condition frame and the synthesized videos, we employ the DINO~\cite{zhang2022dino} visual encoders to extract the visual feature of the given image and of the generated video frames. We then compute the cosine similarity between them to obtain 
% the CLIP First Frame Fidelity Score (FFF$_{CLIP}$) and 
the DINO First Frame Fidelity Score (FFF$_{DINO}$). Regarding the grounding accuracy, which assesses whether the object in the generated video adheres to the specified bounding box sequence, we utilize metrics that are widely used in the context of object tracking~\cite{artrack,trackunifying}, including the Area Overlap (AO) and Success Rate (SR). This process involves employing a pretrained object tracking model, for which we use ARTrack~\cite{artrack}. 

\subsubsection{Image integration strategies} In this work we adhere to the principles of the original ModelScope and GLIGEN, which introduce text and bounding box conditioning through cross-attention and gated self-attention layers, respectively. Our focus is to explore different ways of integrating the condition of the image condition without compromising the effectiveness of the other conditions. We compare with three alternative approaches for integrating the image conditions, including (i) Processing the conditional image through a zero-convolution layer (ZC) and combining it with the noisy latent before inputting to a 3D diffusion UNet, as illustrated in Eq.~\ref{eq:zero_conv}; (ii) Using a similar architecture as ControlNet (CtrlNet)~\cite{zhang2023adding}, with a UNet encoder to process the image condition and then inject this information into the corresponding blocks of the UNet decoder; and (iii) incorporating the image feature into the 3D diffusion UNet via a newly proposed gated cross-attention layer (GCA), as described in Eq.~\ref{eq:cross_gated_attn}. A detailed analysis of these approaches is provided in the following section.

\subsection{Quantitative Results}
\begin{table}[t]
\centering
\caption{Comparison with current methods on the validation split of GOT10K.}
\begin{tabular}{l|ccccc}
\hline 
Methods & FVD($\downarrow$) & FFF$_{DINO}$$(\uparrow)$ & SR$_{50}$$(\uparrow)$\\ \hline 
MS~\cite{wang2023modelscope} & 635& 0.68  & -\\ 
TrackDiff~\cite{tracking} & 563 & - & 0.783 \\
SMCD & \textbf{335} & \textbf{0.85} & \textbf{0.783} \\ 
\hline 
\end{tabular}
\label{fig:sota}
\end{table}
\subsubsection{Comparison with current methods}
To the best of our understanding, our model represents the pioneering effort in enabling video generation based on a combination of image, text, and object motions. Given the absence of prior work that integrates these three signals for video generation, we have opted to benchmark our approach against various methods that incorporate these control signals in distinct manners.

Due to computing limitations, we selected the standard text-to-video model, ModelScope (MS), and TrackDiff, which generates videos based on text and object trajectories, for our benchmarking. Our evaluations are conducted on the validation set of GOT10K.
As demonstrated in Table~\ref{fig:sota}, SMCD markedly surpasses both referenced methods in FVD, highlighting the benefits of integrating image conditions. Furthermore, SMCD not only significantly improves upon MS in terms of FFF$_{DINO}$ scores but also achieves results on SR$_{50}$ that are on par with those of TrackDiff. This underscores our model's robust ability to effectively incorporate multiple input modalities.

\begin{table}[t]
\centering
\caption{Comparison between different image integration strategies on video quality, the results are reported on the validation splits of GOT10K and YTVIS2021.}

\begin{tabularx}{\textwidth}{l*{3}{>{\centering\arraybackslash}X}|*{3}{>{\centering\arraybackslash}X}}
\hline 
\multirow{2}*{Methods}  & \multicolumn{3}{c}{GOT10K} & \multicolumn{3}{c}{YTVIS2021} \\ 
& CLIP-SIM$\uparrow$ & FFF$_{DINO}$$\uparrow$ & FVD$\downarrow$ & CLIP-SIM$\uparrow$ & FFF$_{DINO}$$\uparrow$ &FVD$\downarrow$ \\ \hline  
GT-video & 29.164 & 0.963 & - & 31.491 & 0.950 & -  \\
\arrayrulecolor{lightgray}\midrule[0.25pt]\arrayrulecolor{black}
ZC &28.664&0.851&411&30.931&0.849&413 \\
CtrlNet &28.575&0.843&387&31.080&\textbf{0.851}&368 \\
GCA &\textbf{30.757}&0.731&442&\textbf{32.274}&0.798&498 \\
ZC+CtrlNet &28.232&\textbf{0.852}&421&30.744&0.847&439 \\
% \midrule[0.25pt]\arrayrulecolor{black}
\textbf{ZC+GCA(SMCD)} &29.161&0.851&\textbf{335}&30.767&0.850&\textbf{329}\\ 
\hline 
\end{tabularx}
\label{table:image_integration1}
\end{table}

\begin{table}[t]
\centering
\caption{Comparison between different image integration strategies on grounding accuracy.}
\begin{tabularx}{\textwidth}{l*{3}{>{\centering\arraybackslash}X}|*{3}{>{\centering\arraybackslash}X}}
\hline 
\multirow{2}*{Methods}  & \multicolumn{3}{c}{GOT10K} & \multicolumn{3}{c}{YTVIS2021} \\ 
& SR$_{50}$$\uparrow$ & SR$_{75}$$\uparrow$ & AO$\uparrow$ &  SR$_{50}$$\uparrow$ & SR$_{75}$$\uparrow$  & AO$\uparrow$ \\ \hline  
GT-video & 0.810 & 0.764 & 0.819 & 0.703 & 0.624 & 0.744 \\
\arrayrulecolor{lightgray}\midrule[0.25pt]\arrayrulecolor{black}
ZC &0.731&0.660&0.785&0.657&0.555&0.716 \\ 
GCA &0.702&0.636&0.777&0.644&0.583 &0.726 \\
CtrlNet &0.745&0.702&\textbf{0.798}&\textbf{0.666}&0.592&0.732 \\
ZC+CtrlNet &\textbf{0.788}&0.724&0.679&\textbf{0.666}&0.569&0.727 \\
\textbf{ZC+GCA (SMCD)} &0.780&\textbf{0.728}&0.673&0.646&\textbf{0.593}&\textbf{0.744} \\ 
\hline 
\end{tabularx}

\label{table:image_integration2}
\end{table}

\subsubsection{Image integration strategies}
In this section, we study the performance of different image integration strategies described above, including ZC, CtrlNet and GCA, as well as their combinations. It is worth noting that SMCD adopts the combination of ZC and GCA. 

In Table~\ref{table:image_integration1}, we present the evaluation metrics assessing video quality across the validation splits of GOT10K and YTVIS2021 datasets. Notably, the scores for CLIP-SIM and FFF$_{DINO}$ are consistently high across the various methods evaluated, nearing or even surpassing those of the ground-truth videos (GT-video)\footnote{GT-video serves as a reference rather than the upper bound for these metrics. Since CLIP-SIM evaluates the semantic similarities between texts and videos, achieving a higher score than GT videos is possible.}. However, the FVD scores exhibit significant discrepancies, underlining notable differences in video quality among the various approaches. Furthermore, Table~\ref{table:image_integration2} introduces the evaluation metrics for grounding accuracy. This assessment aims to determine the compatibility of these methods with the pretrained motion integration module, evaluating their effectiveness in accurately generating objects with location guidance.

% in both video quality and grounding accuracy
Interestingly, employing a single CtrlNet module yields better FVD compared to both ZC and GCA individually. However, when ZC is combined with CtrlNet, the performance on FVD deteriorates for both datasets. This decline could be attributed to the methodological overlap where ZC and CtrlNet: both enhance image conditions through zero-convolutional layers and feature addition. Specifically, ZC integrates the conditional image with a noised image for the encoder, while CtrlNet introduces the encoded clean image to the decoder. This dual approach to processing the conditional image might lead to conflicts, adversely affecting the outcome.

Conversely, integrating ZC and GCA within SMCD framework leads to significant enhancements.  It achieves the best results in video quality and secures superior outcomes in grounding accuracy, which is even close to that of ground-truth videos.
This synergy, driven by a flexible gate function that modulates the intensity of image signals, ensures that these two components work in harmony, resulting in a performance that surpasses that of other strategies explored. 

\begin{table}[t]
\centering
\caption{Ablation on the impact of each component in SMCD on GOT10K val. set.}
\renewcommand{\arraystretch}{1.2} % Increase the row height
\begin{tabular}{l|ccc|cccc}
\hline
Methods & Txt & Img & Box & FVD($\downarrow$) & CLIP-SIM($\uparrow$) & FFF$_{DINO}$$(\uparrow)$ & SR$_{50}(\uparrow)$ \\ \hline
MS & \checkmark & - & - & 635 & 29.10 & 0.68 & - \\
% MS+MIM & \checkmark & - & \checkmark & 743 & 27.88 & 0.69 & $0.58$/$0.48$/$0.69$ \\
% MS+DIIM & \checkmark & \checkmark & - & 353 & 28.23 & 0.83 &$0.52$/$0.40$/$0.65$  \\
MS+MIM & \checkmark & - & \checkmark & 425 & 28.01 & 0.71 & 0.62 \\
MS+DIIM & \checkmark & \checkmark & - & 388 & 28.91 & 0.84 & 0.57  \\
SMCD & \checkmark & \checkmark & \checkmark & \textbf{335} & \textbf{29.16} & \textbf{0.85} & \textbf{0.78} \\
\hline
\end{tabular}
\label{tbl:add-on-ablation}
\end{table}

\subsection{Ablative Studies}
% \subsubsection{Ablation on each component in SMCD}
In this section, we delve into the significance of each component within the SMCD framework, where we incrementally introduce the motion integration module (MIM) and the dual image integration module (DIIM) to the foundational MS model. The performance enhancements brought about by these additions are detailed in Table~\ref{tbl:add-on-ablation}.

% For simplicity, these results are obtained by designating the image input $v_0$ or box trajectory input $O$ as \textit{null} during inference.

\begin{figure}
    \centering
    \includegraphics[width=1.0\linewidth]{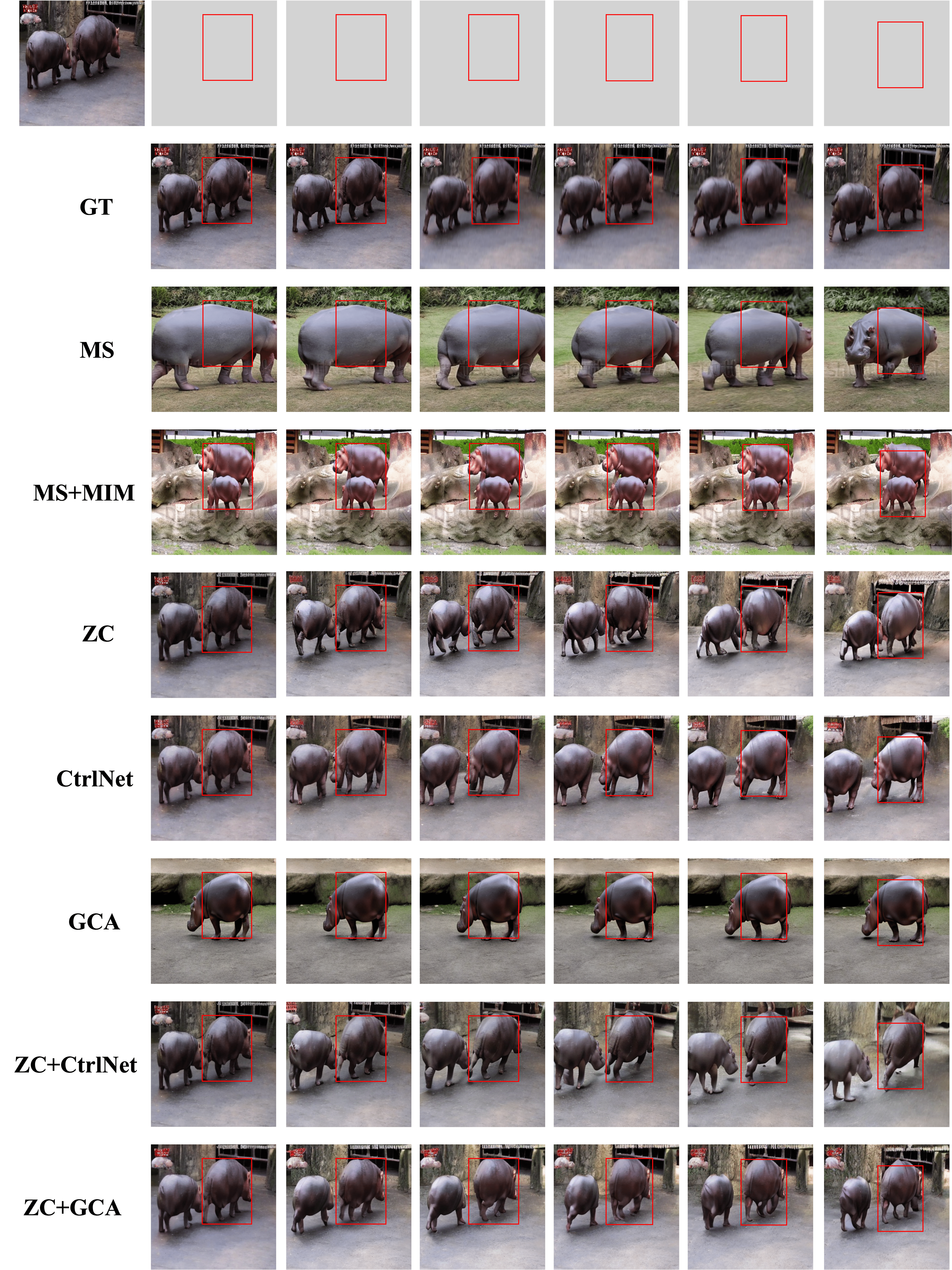}
    \caption{\textbf{Qualitative comparison of videos generated by different models.} The given caption is: \textit{A hippopotamus that is walking}. Our dual image integration module, labeled ZC+GCA, excels in maintaining the complex semantics of the initial frame while precisely conforming to the motion dynamics defined by the sequence of provided bounding boxes. Additionally, it ensures the retention of high video quality.
    }
    \label{fig:visual-3}
\end{figure}

Integrating MIM alone was observed to detrimentally affect the FVD and CLIP-SIM scores. This decline can be attributed to the motion control counteracting MS's original generative intent, compelling the model to localize objects in specific regions, as evidenced by Fig.~\ref{fig:visual-3}.
Furthermore, incorporating DIIM alone leads to significant improvements in video quality including FVD and FFF$_{DINO}$. Interestingly, it yields reasonable grounding metrics, attributable to the conditional image furnishing a prior for object locations. However, in the absence of motion guidance, these metrics remain on the lower side. The combination of both modules within SMCD leads to optimal outcomes, significantly boosting both semantic consistency and motion control.

% \subsubsection{Gated Cross-attention vs Parallel cross-attention}
% \begin{table}[h]%\scriptsize
% \centering
% \begin{tabular}{l|cccccc}
% \hline 
% \multirow{2}*{Methods}  & \multicolumn{6}{c}{GOT10K} \\ 
% ~& CLIP-SIM$\uparrow$ & FFF$_{DINO}$$\uparrow$ & FVD$\downarrow$ &  SR$_{50}$$\uparrow$ & SR$_{75}$$\uparrow$  & AO$\uparrow$ \\ \hline  
% cross &0.731&0.660&0.785&0.657&0.555&0.716 \\ 
% cross-parallel &0.702&0.636&0.777&0.644&0.583 &0.726 \\
% %cross-parallel &0.664&0.599&0.756&?&? \\
% cat+cross &0.745&0.702&0.798&0.666&0.592&0.732 \\
% cat+cross-parallel &0.788&0.724&0.679&0.666&0.569&0.727 \\ 
% \hline 
% \end{tabular}
% \caption{Example of a 3x6 table.}
% \label{table:3x6}
% \end{table}

\subsection{Qualitative Results}
We present some qualitative results for an intuitive comparison between different methods. As shown in Figure~\ref{fig:visual-3}, the inputs are the conditional image, the moving trajectory, and the text prompt ``\textit{A hippopotamus that is walking}''.
The ModelScope model (MS) generates a video aligned with the text but without considering any motion or image conditions. Equipped with MIM, it can generate a hippopotamus in the corresponding region but in a different scene. Various image integration methods enable models to replicate the condition image's frames closely. However, the GCA focuses on high-level semantic similarities, missing finer details. CtrlNet induces rapid movements, exemplified by a hippopotamus's head shifting swiftly from right to left. The ZC method effectively maintains semantic integrity and adheres to the motion trajectory but falls short in preserving subject consistency, as illustrated by the color change of the small hippopotamus, which turns white in the final frame. Similar to the ZC, the combination of ZC and CtrlNet generates a video that accurately captures the given conditions, but fails at keeping background consistency. Finally, our SMCD method (ZC+GCA) generates a video that not only meets the specified conditions but also maintains consistency across frames. For more qualitative results, we refer to supplemental material.

\section{Conclusion}
In this study, we propose augmenting a text-to-video generation model with additional control signals, thereby generating videos that better align with human intentions. To this end, we investigate various methods for incorporating diverse control signals into the diffusion UNet framework and introduce the SMCD (Scene and Motion Conditional Diffusion Model). At the heart of our design are the MIM (Motion Integration Module) and DIIM (Dual Image Integration Module) modules, which are meticulously crafted to seamlessly integrate image and bounding box sequences, into the diffusion U-Net. Experimental results indicate that our method effectively integrates control signals into a diffusion UNet, generating high-quality videos that align closely with the provided control signals.

\textbf{Limitation and Future Works} While our method successfully generates high-quality videos that meet three specified conditions, we find that relying solely on box sequence conditions is insufficient for controlling object dynamics, as similar changes can result from camera movement. For comprehensive control, incorporating camera constraints is crucial, which we plan to explore in future research. 
Additionally, we have observed that some videos exhibit shaking, likely due to significant camera movement in the training data and a reduced frame rate during training. Potential solutions include training with a higher sampling rate and incorporating camera pose to distinguish between camera and object movements, which will be considered in future research. Lastly, using the pretrained ModelScope model as our backbone inherits its constraints, including difficulties in generating high-quality videos of humans.

% \clearpage\mbox{}Page \thepage\ of the manuscript.
% \clearpage\mbox{}Page \thepage\ of the manuscript.
% \clearpage\mbox{}Page \thepage\ of the manuscript.
% \clearpage\mbox{}Page \thepage\ of the manuscript.
% \clearpage\mbox{}Page \thepage\ of the manuscript. This is the last page.
% Now we have reached the maximum length of an ECCV \ECCVyear{} submission (excluding references and acknowledgements).
% References should start immediately after the main text, but can continue past p.\ 14 if needed. 
% \clearpage  % TODO FINAL: This \clearpage needs to be removed from both review and camera-ready versions.

\section*{Acknowledgements}
This work is funded by the European Research Council under the European Union’s Horizon 2020 research and innovation program
(projects CALCULUS, H2020-ERC-2017-ADG 788506 and KeepOnLearning, H2020-ERC-2021-ADG 101021347) and the Flanders AI Research Program. We also acknowledge the valuable discussion with Pengxiang Li.

\bibliographystyle{splncs04}
\bibliography{main}
\include{appendix_for_arxiv}
\end{document}

%% file: appendix_for_arxiv.tex
\title{Supplementary material of Animate Your Motion: Turning Still Images into Dynamic Videos}

\titlerunning{Animate Your Motion}

% Second Author\inst{2,3}\orcidlink{1111-2222-3333-4444} \and
% Third Author\inst{3}\orcidlink{2222--3333-4444-5555}}

% for arxiv
\author{ Mingxiao Li$^*$\inst{1} \and
 Bo Wan$^*$\inst{2} \and
Marie-Francine Moens\inst{1} \and 
Tinne Tuytelaars\inst{2}}

\authorrunning{Li et al.}

\institute{Department of Computer Science, KU Leuven
\email{\{mingxiao.li,sien.moens\}@kuleuven.be}\\
%\url{http://www.springer.com/gp/computer-science/lncs} 
\and
Department of Electrical Engineering, KU Leuven\\
\email{\{bo.wan,Tinne.Tuytelaars\}@esat.kuleuven.be} \\
}

\maketitle
\def\thefootnote{*}\footnotetext{Equal Contribution. Project page: \href{https://mingxiao-li.github.io/smcd/}{https://mingxiao-li.github.io/smcd/}}\def\thefootnote{\arabic{footnote}}

\section{More visualization}
\begin{figure}[h]
    \centering
    \includegraphics[width=1.0\linewidth]{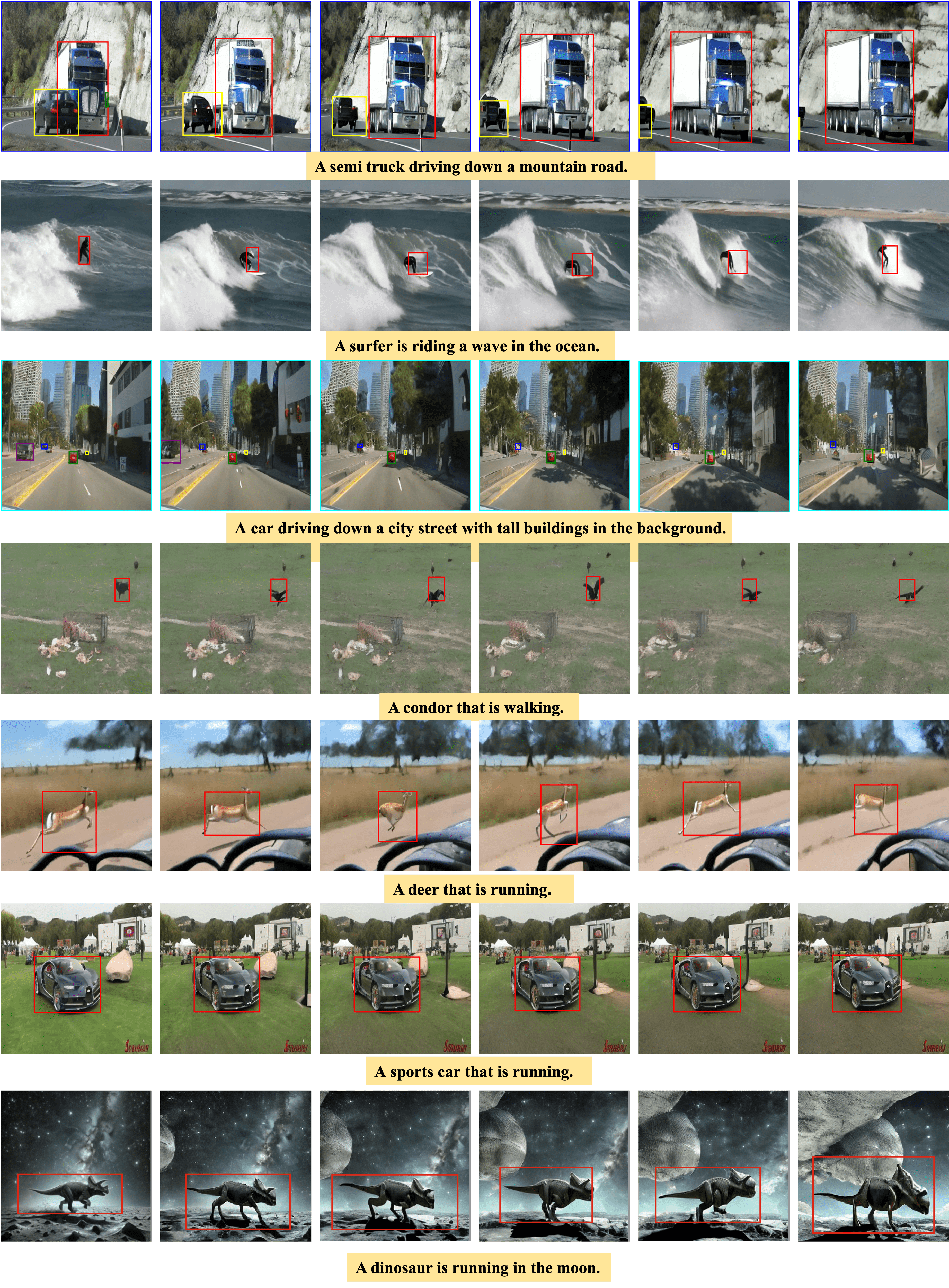}
    \caption{\textbf{Scene motion customized video generation results} of our proposed SMCD model. }
    \label{fig:visual-4}
\end{figure}

In this section, we showcase additional videos generated by SMCD as depicted in Fig.~\ref{fig:visual-4}. These results illustrate SMCD's ability to produce videos by integrating both image and motion conditions. 
As SMCD uses the CLIP text encoder for object semantic embedding, 
and learns general knowledge to maintain semantic and motion consistency, 
excelling with out-of-distribution categories. It performs well in tests with novel scenarios like ‘A dinosaur running in the moon’, as shown in the last row of Fig.~\ref{fig:visual-4}.

% Additionally, we provide the videos in the supplementary materials. 
% 0a9e0643c3  029c028307  020cd28cd2 00703ca71a  GOT-10k_Val_000031  GOT-10k_Val_000069 GOT-10k_Val_000088 GOT-10k_Val_000125 GOT-10k_Val_000169
% Imaginary
% GOT-10k_Val_000004
% GOT-10k_Val_000039

\section{Comparison between different approaches} 
\begin{figure}[ht]
    \centering
    \includegraphics[width=1.0\linewidth]{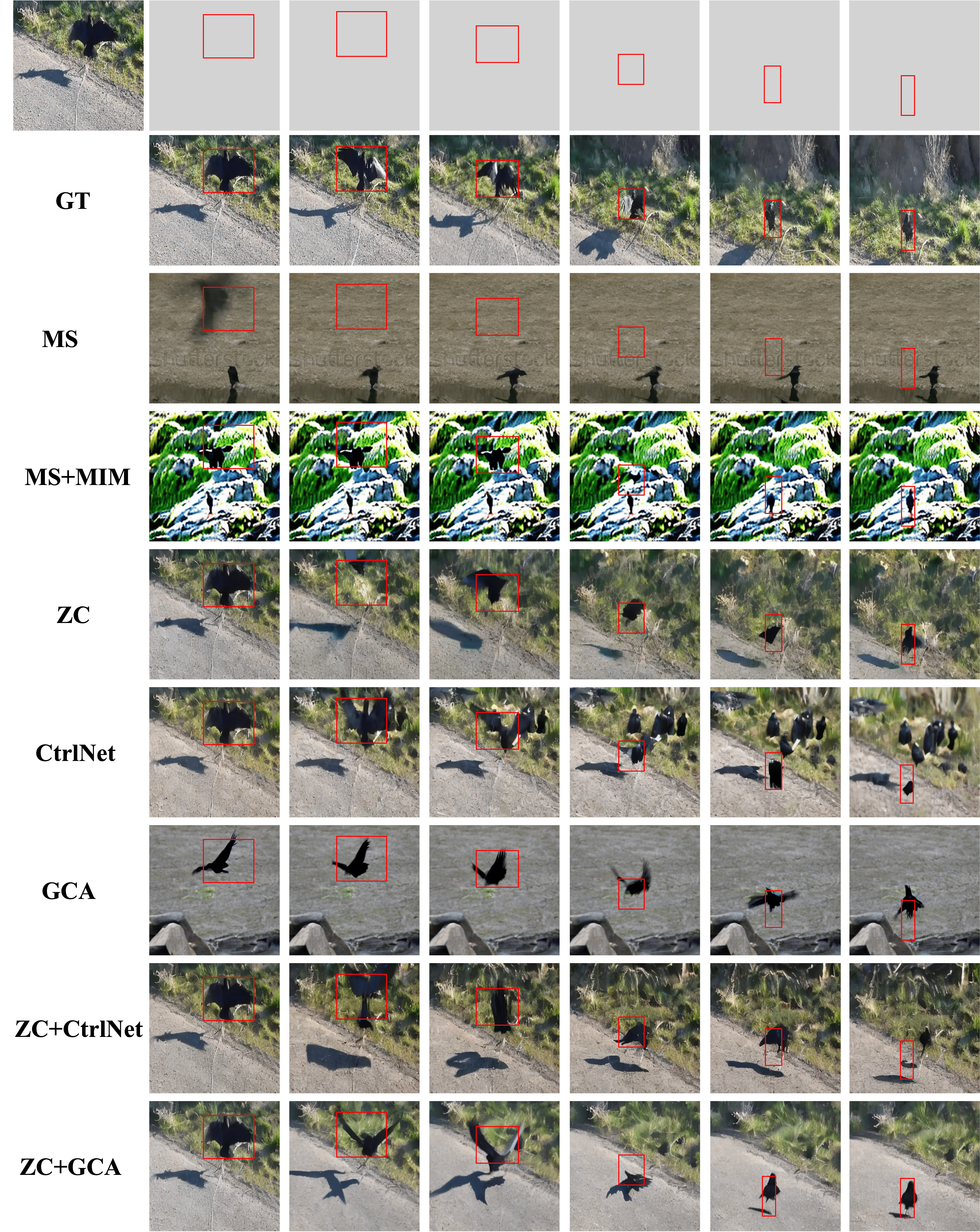}
    \caption{\textbf{Qualitative comparison of videos generated by different models.} The given caption is: \textit{A hippopotamus that is walking}.}
    \label{fig:compare-e}
\end{figure}

In Fig.~\ref{fig:compare-e}, we offer more qualitative comparisons among various models, similar to Fig.~\textcolor{red}{3} in the main paper. Similarly, we find ModelSopde (MS) fails to accurately represent the dynamics of the bounding box sequence. Additionally, both the MS+MIM and GCA models fail to preserve the semantic content of the initial frame. Conversely, ZC, CtrlNet and ZC+CtrNet are challenged by issues related to poor visual quality. Our proposed SMCD model stands out by successfully generating high-quality videos that not only retain the semantic integrity of the first frame but also accurately follow the motion described by the bounding box sequence.

\section{Failure cases} 
Despite the superior performance achieved by SMCD, it still encounters some limitations, which we highlight through two typical failure cases in this section. 

\begin{figure}[ht]
    \centering
    \includegraphics[width=1.0\linewidth]{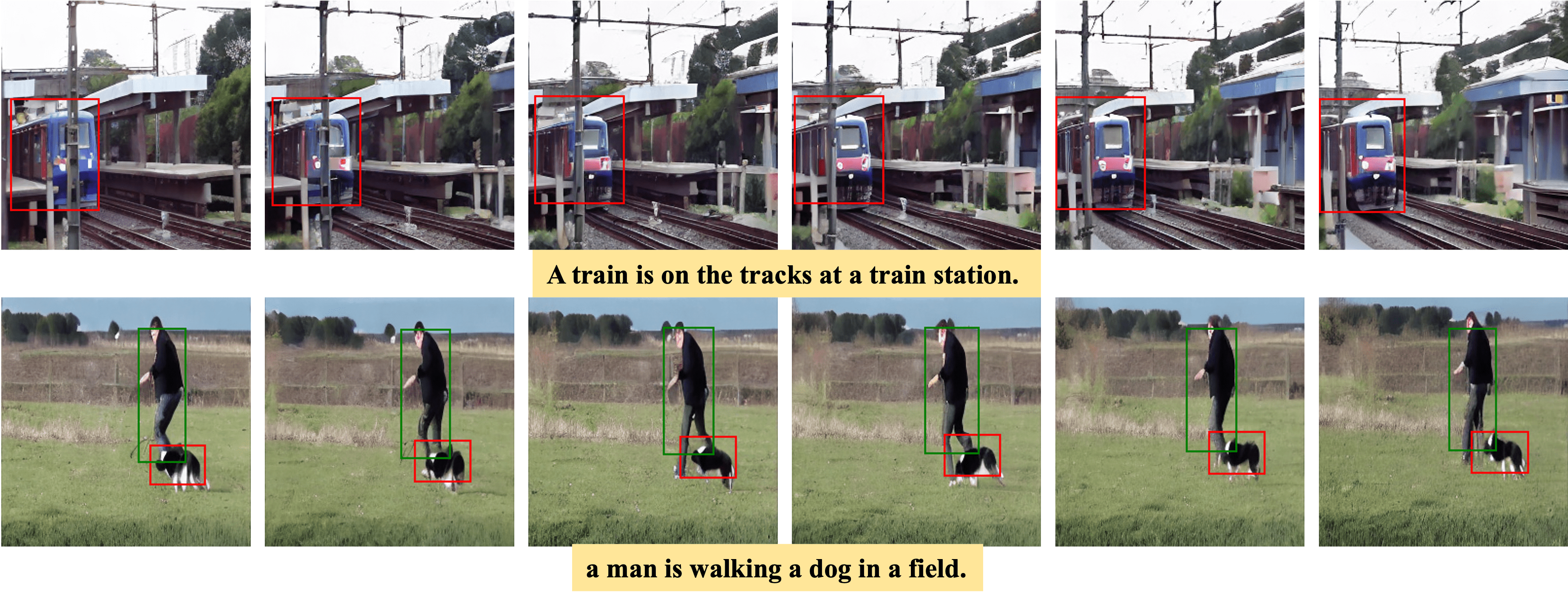}
    \caption{Failure cases of color change between video frames.}
    \label{fig:visual-color}
\end{figure}

The first issue is the inconsistency in object colors across different frames, as illustrated in Fig.~\ref{fig:visual-color}. This inconsistency arises because the model attempts to transfer colors from one region to another. For instance, in the first example of Fig.~\ref{fig:visual-color}, the train's head starts off blue while its body is red, and over time, the model gradually changes the color of the train's head to red. This subtle shift can be attributed to the self-attention layers in the diffusion UNet, which facilitate the spread of semantic information within an object.

\begin{figure}[ht]
    \centering
    \includegraphics[width=1.0\linewidth]{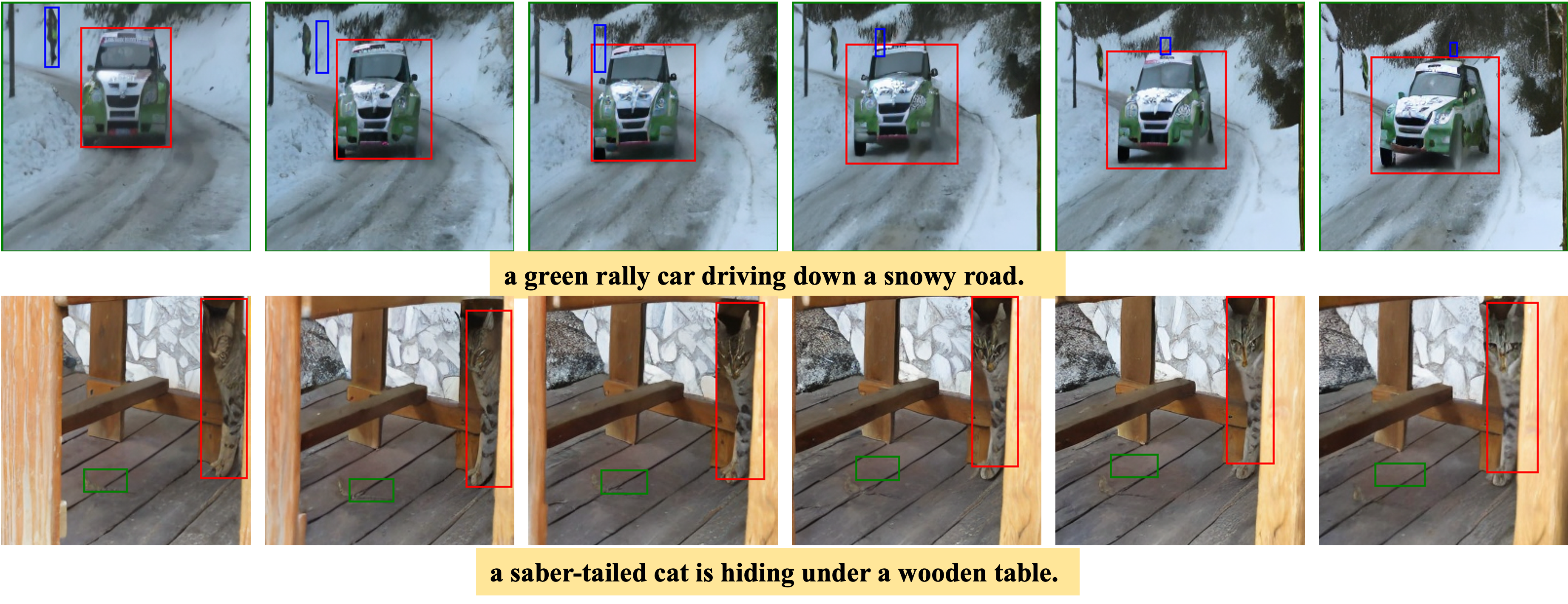}
    \caption{Failure cases of overlooking small objects during video generation.}
    \label{fig:visual-small}
\end{figure}

The second challenge involves the model's oversight of small objects during generation, as depicted in Fig.~\ref{fig:visual-small}. This problem arises when the model is compelled to create small objects in specific areas, despite the pretrained video diffusion model (ModelScope) not being equipped with the capability to do so effectively.

\section{More ablative studies}
\paragraph{Ablation on training scheme}
Our two-stage training pipeline is detailed in Sec.~\textcolor{red}{3.4} in the main paper. 
To provide a comprehensive analysis, we also explored the performance of a 
joint training approach, as illustrated in the second row of Tab.~\ref{tbl-abl}. The results from this comparison indicate that the joint training methodology yielded suboptimal outcomes when contrasted with the two-stage training pipeline.

\paragraph{Ablation on model design}
In this ablative study, we examined the influence of layer orders on system performance. 
To maintain the integrity of our experimental framework, we adhered to the original ModelScope layer order. 
Additionally, we conducted test to evaluate the effects of altering the sequence of the newly incorporated object and image attention layers. 
As shwon the third row in Tab.~\ref{tbl-abl}, modifying the order of these added layers had minimal impact on the overall performance of the system. This suggests that the positional arrangement of the object and image attention layers within the network architecture is not a critical factor in determining the model's effectiveness.

\begin{table}[t]
    \centering
    \caption{Ablation on training scheme and model design.}
    \begin{tabularx}{\textwidth}{X|XXXX}
    \hline 
     Methods & FVD $\downarrow$ & CLIP-SIM $\uparrow$ & FFF$_{DINO}$ $\uparrow$ & SR$_{50}$ $\uparrow$ \\ \hline 
     SMCD & \textbf{335} & \textbf{29.16} & \textbf{0.85} & \textbf{0.78} \\ 
     Joint & 385 & 28.71 & 0.83 & 0.69 \\
     Swap & 340 & 28.98 & 0.84 & 0.76 \\
    \hline
    \end{tabularx}
    \label{tbl-abl}
\end{table}

%Videos of Fig. 3
%One or two more examples and their videos

%\section{Failure cases}
% color change: 058a7592c8  029c028307  0630391881
% ignore small objects: 078b9fff92 0a7a2514aa

\clearpage  % TODO REVIEW/FINAL: This \clearpage needs to be removed from both review and camera-ready versions.

% \bibliography{egbib}